# Identifying causal relations in tweets using deep learning: Use case on diabetes-related tweets from 2017-2021

## Authors

Adrian Ahne[1,2], Vivek Khetan[3], Xavier Tannier[4], Md Imbesat Hassan Rizvi[5], Thomas Czernichow[2], Francisco Orchard[2], Charline Bour[6], Andrew Fano[3], Guy Fagherazzi[6]

1. Paris-Saclay University, UVSQ, Inserm, Gustave Roussy, "Exposome and Heredity" team, CESP, F-94805, Villejuif, France
2. Epiconcept Company, Paris, France
3. Accenture Labs, San Francisco, USA
4. Sorbonne University, Inserm, University Sorbonne Paris Nord, Laboratoire d'Informatique Médicale et d'Ingénierie des Connaissances pour la e-Santé, LIMICS, Paris, France
5. Indian Institute of Science, Bengaluru, India
6. Deep Digital Phenotyping Research Unit, Department of Precision Health, Luxembourg Institute of Health, Strassen, Luxembourg

**Corresponding author:** Adrian Ahne,

Paris-Saclay University, UVSQ, Inserm, Gustave Roussy, "Exposome and Heredity" team, CESP - Epiconcept, 20 rue du Dr. Pinel, 94800 Villejuif, France
Phone: +33 684 70 53 39
Email: a.ahne@epiconcept.fr

**Keywords:** Causality, deep learning, natural language processing, Diabetes, Social media

# Abstract

**Background**

Intervening and preventing diabetes distress requires an understanding of its causes and in particular from a patients' perspective. Social media data provide direct access to how patients see and understand their disease and in consequence express causes of diabetes distress.

**Objective**

Leveraging machine learning methods, we aim to extract both explicit and implicit *cause-effect* relationships in patient-reported, diabetes-related tweets and provide a methodology to better understand opinion, feelings and observations shared within the diabetes online community from a causality perspective.

**Methods**

More than 30 million diabetes-related tweets in English were collected between April 2017 and January 2021. Deep learning and natural language processing methods were applied to focus on tweets with personal and emotional content. A *cause-effect-tweet* dataset was manually labeled and used to train 1)  a fine-tuned Bertweet BERTweet model to detect *causal sentences* containing a causal causal relation; 2) a Conditional Random Field (CRF) model with BERT based features to extract possible cause-effect associations. Causes and effects were clustered in a semi-supervised approach and visualised in an interactive cause-effect-network.

**Results**

*Causal sentences* were detected with a recall of 68% in an imbalanced dataset. A CRF model with BERT based features outperformed a fine-tuned BERT model for *cause-effect* detection with a macro recall of 68%. This led to 96,676 sentences with cause-effect relationships. "Diabetes" was identified as the central cluster followed by "Death" and "Insulin". Insulin pricing related causes were frequently associated with "Death".

**Conclusions**

A novel methodology was developed to detect causal sentences and identify both explicit and implicit, single and multi-word cause and corresponding effect as expressed in diabetes-related tweets leveraging BERT-based architectures and visualised as cause-effect-network. Extracting causal associations on real-life, patient reported outcomes in social media data provides a useful complementary source of information in diabetes research.

# Introduction

Diabetes distress (DD) refers to psychological factors such as emotional burden, worries, frustration or stress in the day-to-day management of diabetes.[1–3] Diabetes distress is associated with poorer quality of life,[4] higher A1C levels[5(p),6] and medication adherence.[7] Reducing diabetes distress may improve hemoglobin A1c and reduce the burden of disease among people with diabetes.[8] Social media is a useful observatory resource for patient reported diabetes issues and could help to contribute directly to public and clinical decision making from a patients' perspective, given the active online diabetes community.[9,10]
Identifying causal relations in social media expressed text data might help to discover unknown etiological results, specifically causes of health problems, concerns and symptoms. To intervene and potentially prevent diabetes distress it is necessary to understand the causes of diabetes distress from a patients' perspective to understand how patients see their disease. Causal relation extraction in natural language text has gained popularity in clinical decision-making, biomedical knowledge discovery or emergency management.[11] Particularly, causal relations on Twitter have been examined for diverse factors causing stress and relaxation,[12] adverse drug reactions [13] or causal associations related to insomnia or headache.[14]

In this paper, we aim to extract spans of text as two distinct events from diabetes and diabetes-related tweets such that one event directly or indirectly impacts another event. We categorized these events as cause-event and effect-event depending upon the expressed context of each tweet. This work is realised in the frame of the World Diabetes Distress Study (WDDS) which aims to analyze what is shared on social media worldwide to better understand what people with diabetes and diabetes distress are experiencing.[15,16]

Most approaches examine *explicit* causality in text,[14,17,18] when cause and effect are explicitly stated, for instance by connective words (e.g. so, hence, because, lead to, since, if-then).[11,19] An example for an *explicit* cause-effect pair is “diabetes causes hypoglycemia”. Whereas, *implicit* causality is more complicated to detect such as in “I reversed diabetes with lifestyle changes” with cause “lifestyle changes” and effect “reversed diabetes”.
Machine and deep learning models have also been applied to extract causal relations. They are able to explore implicit relations and provide better generalisation contrary to rule-based approaches.[11,20,21,22(p)] An interesting approach, leveraging the transfer learning paradigm, and addressing both explicit and implicit cause-effect extraction is provided by Khetan et al.[23] They fine-tuned pre-trained transformer based BERT language models[24,25] to detect “Cause-Effect” relationships using publicly available datasets such as the adverse drug effect dataset.[26]

In a similar spirit, the objective of the present work is to identify both explicit and implicit multi-word cause-effect relations on noisy, diabetes-related tweets, to aggregate identified causes and effects in clusters and ultimately to visualise these clusters in an interactive cause-effect network.

# Materials and Methods

On the basis of diabetes-related tweets, we first preprocessed tweets to only focus on personal, no-joke and emotional content; secondly, we identified tweets in which causal information (opinion, observation, etc.) is communicated, also referred to as causal tweets or causal sentences; in a third step, causes and their corresponding effects were extracted. Lastly, those cause-effect pairs were aggregated, described and visualised. The entire workflow is illustrated in Figure 1.

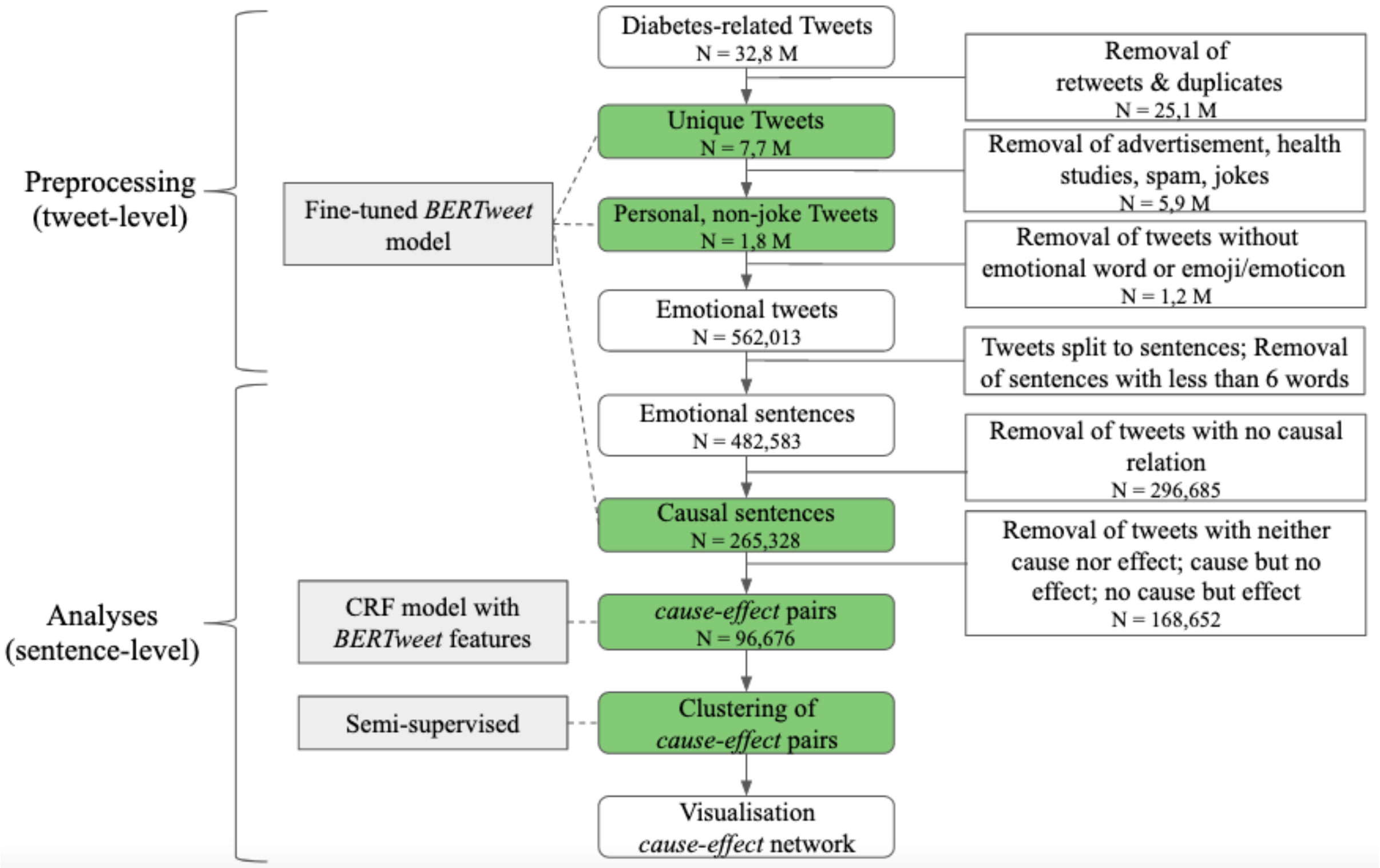

Figure 1: Workflow. In green color the steps including machine learning methods

# Data collection

Via Twitter's Streaming Application Programming Interface (API), 32 million diabetes-related tweets in English were collected between April 2017 and January 2021 based on a list of diabetes-related keywords, such as *diabetes, hypoglycemia, hyperglycemia* and *insulin*, from all over the world (see Multimedia Appendix 1 for the full list of keywords used). This is an extended dataset of the one used in earlier works.[9] All data collected in this study were publicly posted on Twitter. Therefore, according to the privacy policy of Twitter, users agree to have this information available to the general public.[27]

# Data preprocessing

In this work, we applied a preprocessing pipeline, similar to earlier works,[9] to focus on tweets with personal content and remove institutional tweets (organisations, advertisement, news, etc.), identify and exclude jokes, and filter tweets containing emotional elements to adjust the scope of the tweets towards diabetes distress. Besides, questions were removed. This led to 562,013 tweets containing personal, non-joke and emotional content. More details on the preprocessing pipeline are summarized in the Multimedia Appendix 2.

## Data annotation

In order to identify causal tweets and *cause-effect* association, 5,000 randomly chosen diabetes-related tweets were manually labeled. We did not restrict ourselves to a specific area of diabetes-related causal relationships and include potentially all types.

Table 1 illustrates some example tweets. For a more detailed explanation on the annotation please refer to our annotation guidelines in Multimedia Appendix 3.

| Tweet | Cause | Effect | C.A.* | Explanation |
|---|---|---|---|---|
| Diabetes causes me to have mood swings | Diabetes | mood swings | 1 | **Possible causal association** |
| I just want to eat . I hate #diabetes | #diabetes | hate | 1 | **Possible causal association** related to diabetes distress |
| Scary, have a diabetic daughter but I read thousands of people a year die in the UK just from flu so why panic over corona . | | | 0 | **Non-diabetes or diabetes distress related relationship.** “flu” is not diabetes-related |
| I'm back ! Had two strokes and recover now. Have high blood pressure and diabetes. :-) | | | 0 | **Unclear cause-effect relationship** Not clear if “High blood pressure” or “diabetes” caused the stroke |
| Not sure if I've been up since 3:30 to watch Titanic or because of my anxiety over my glucose test is what keeps me up 😬 | glucose test | anxiety | 1 | **Chaining cause-effect relationship A->B->C**<br>event A: glucose test<br>event B: anxiety<br>event C: been up since 3:30<br>=> label the relationship which is closest to our study objective: diabetes and diabetes distress |
| My 14 year old daughter is Type 1 = malfunctioning pancreas , meaning not enough insulin being made to regulate 😟 | Type 1 | malfunctioning pancreas; not enough insulin | 1 | **Negation** negation in a cause/effect is considered being part of the cause/effect as it does not alter the meaning |
| it is not true to think that insulin makes you feel so bad 🙃 | insulin | feel so bad | 0 | **Negation** negation is not part of cause/effect and alters the meaning |

Table 1: Sample tweets in different label scenarios. The tweets are fictive to ensure privacy but represent similar real tweets

*C.A.: causal association

Labeling cause-effect pairs is a complex task. To verify the reliability of the labeling, two authors labeled 500 tweets independently and we calculated Cohen's kappa score, a statistical measure expressing the level of agreement between two annotators.[28] We obtained a score of 0.83, which is interpreted as an *almost perfect* agreement according to Altman and Landis.[29,30] Disagreements were discussed between the two authors and one author labelled additional 4,500 tweets, resulting in 5,000 labeled tweets.

## Models

A first model was trained to predict if a sentence contains a potential cause-effect association (causal sentence) and a second model extracted the specific cause and associated effect from the causal sentence. Thus, the first model acts like a barrier and filters non-causal sentences out. These tweets may have either a cause, an effect, none of them, but not both. To simplify the model training, we hypothesised that cause-effect-pairs only occur in the same sentence and we removed all sentences with less than 6 words due to a lack of context. For this reason we operated on a sentence and not tweet level.

Additional challenges in our setting were that *causes* and *effects* could be multi word entities and the language used on Twitter is non-standard with frequent slang and misspelled words.

### Causal sentences detection

The identification of causal sentences is a binary classification task. The pre-trained language model *BERTweet* served as foundation for the model architecture capable of handling the non-standard nature of Twitter data.[31] A feed-forward network is built on top of the *BERTweet*[31] architecture consisting of two fully connected layers (FCLL) with dropout layers with probability 0.3, finalized by a softmax layer which translates the model predictions into probabilities, see Figure 2. The initial training set of 5,000 tweets was imbalanced and, after splitting the tweets into sentences, resulted in 7,218 non causal sentences and 1,017 causal sentences. To adjust for the data imbalance, class weights were included as parameters in the categorical cross entropy loss function to penalise mispredictions for causal sentences stronger. Parameters for the model training were the adam optimizer with a learning rate of 1e-3, *epsilon* of 1e-8, a scheduler with linearly decreasing learning rate and 0 warm up steps, learning rate of 1e-3 and we trained for 35 epochs with early stopping. Initially labelled data was stratified and 10% of it was kept as test set. The remaining 90% samples were further separated into training and validation set with 80:20% split. Batch size for training and validation was 16 and 32 for the test set.

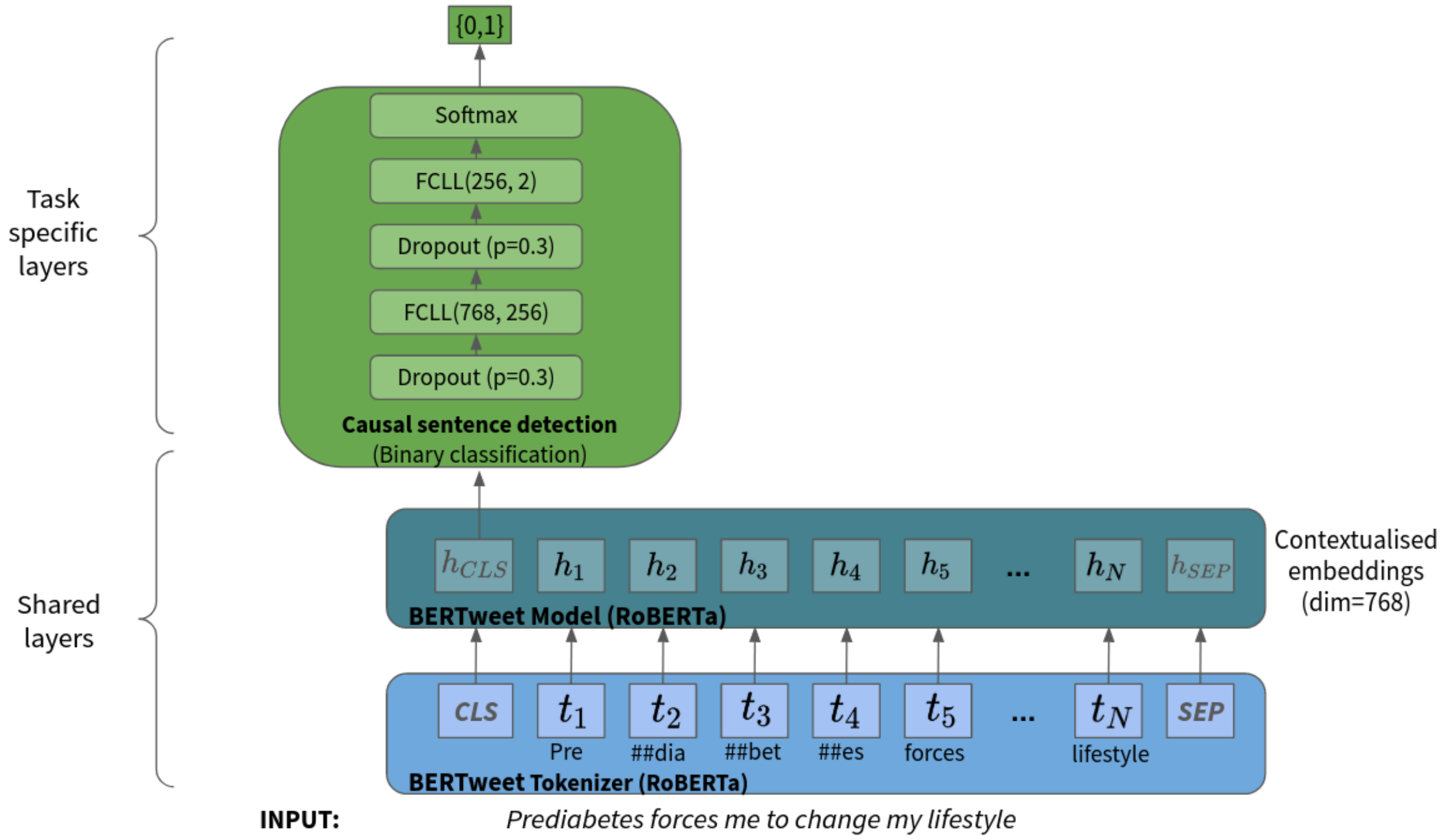

Figure 2: Model architecture - Causal sentence detection

## Data augmentation through active learning

Data imbalance on the one hand and on the other hand the limited number of positive training examples for each cause-effect pair, due to the fact that causes and effects could potentially be related to any concept in the diabetes domain, drove us to adopt an active learning approach to increase the training data. Active learning is a sample selection approach aiming to minimize the annotation cost while maximising the performance of ML-based models.[32] It has been widely applied on textual data.[33,34] The training data was increased in several iterations as illustrated in Figure 3.

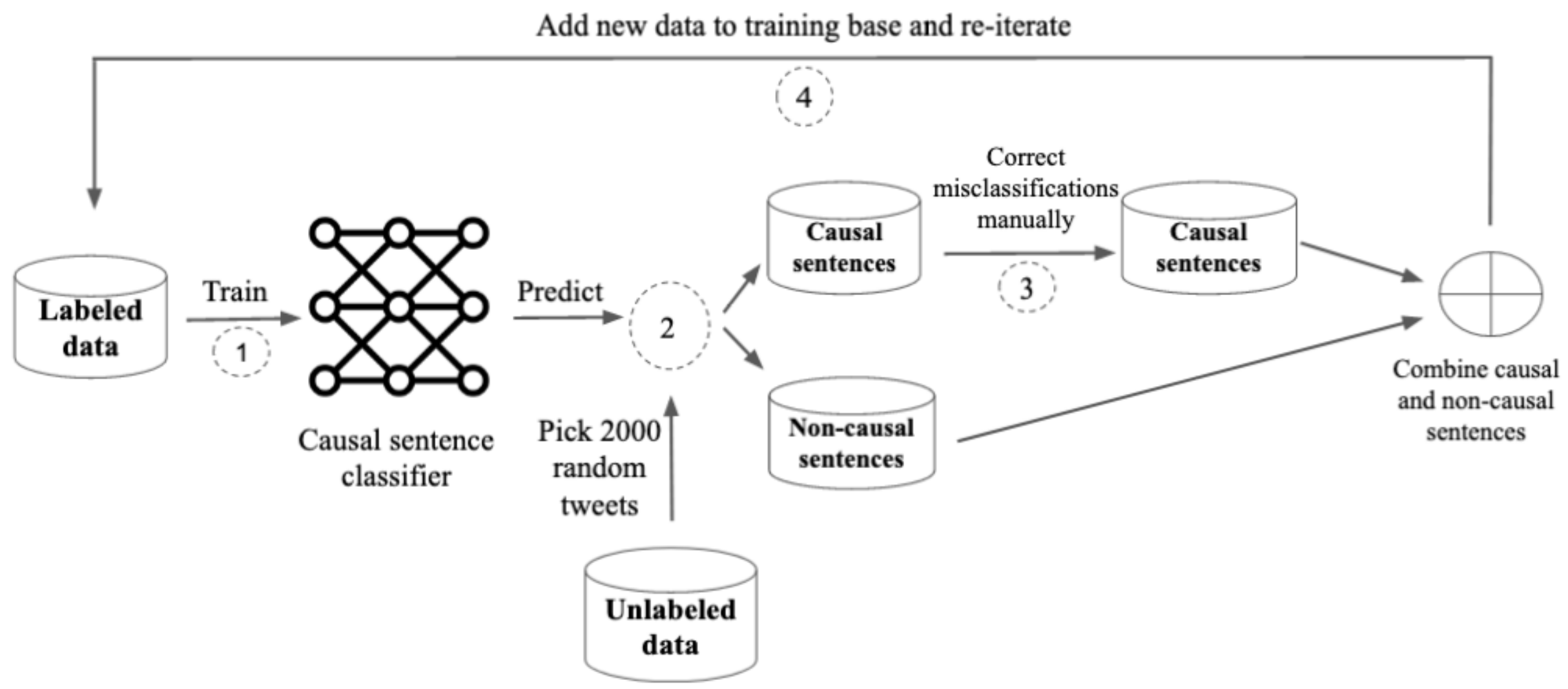

Figure 3: Active learning loop to augment the training set in a time-efficient fashion

The first iteration started by training the causal sentence classifier on sentences from the 5,000 tweets. The trained classifier was then applied on 2,000 randomly selected unlabeled tweets resulting in a set of causal sentences and a set of non-causal sentences. The sentences predicted as causal sentences were examined manually and possible misclassifications were corrected to ensure clean positive training samples. The non-causal sentence set remained untouched. As a consequence, potential misclassifications remained in the non-causal sentence set, which should then be considered noisy. Both the causal and non-causal sentence set were then combined and added as new training data to the already labeled data, leading to an updated training set of 7,000 tweets. This process was iterated four times, and allowed us to augment the labelled data much faster and efficiently than without active learning as it enables us to focus on the few positive samples. The final training set was used to train the classification model and the cause-effect extraction model.

## Cause-effect pairs

After having trained the causal sentence classifier to detect sentences with causal information, we identified the specific cause-effect pairs in the causal sentences. The identification of cause-effect pairs was casted as an event extraction, or named entity recognition task, i.e assigning a label cause or effect to a sequence of words. The manually labeled *causes* and *effects* were encoded in a IO tagging format based on the common tagging format BIO (Beginning, Inside, Outside), introduced by Ramshaw and Marcus.[35] Here, “I-C” denotes inside the cause and "I-E" inside the effect. Those two tags were completed by the outside tag "O" symbolizing that the word is neither cause nor effect. The IO tagging scheme for the example sentence with cause *Prediabetes* and effect *change my lifestyle* is summarized:

| | | | | | | | |
|---|---|---|---|---|---|---|---|
| Sentence : | Prediabetes, | forces, | me, | to, | change, | my, | lifestyle |
| IO tags: | I-C | O | O | O | I-E | I-E | I-E |

Note that a word can be both cause or effect depending on the context. For instance "*Prediabetes*" in "*Prediabetes forces me to change my lifestyle"* takes the role of a *cause*, whereas in "*Limited exercising may lead to prediabetes"* it is a possible *effect*. IO-tagging was preferred over BIO-tagging to simplify the model learning by reducing the number of class from five to three. Moreover, the task is complex and considered open-domain as *causes* and *effects* are not restricted to one specific topic, but can be related to any concept in our target domain (diabetes). As a consequence, the creation of a representative training set is challenging, as most *cause-effect* pairs occur rarely.

This complexity drove us to test several model architectures, compare Figure 4 for an overview:

1. **BERT_FFL:** Pre-trained *BERTweet* language model and on top two feed forward layers with a dropout of 0.3, followed by a softmax layer. For the model training the cross entropy loss function is selected and weighted by the class weights to penalise mispredictions for causes and effects stronger.

2. **WE_BERT_CRF:** Single CRF layer with *BERTweet* embeddings as features augmented by discrete features such as if the word is lowercase, digit or the word length. The CRF function is implemented with the python package *sklearn-crfsuite*[37] based on *CRFsuite.*[38] As parameters for the CRF function, the default algorithm "Gradient descent using the L-BFGS method" was chosen and coefficients for L1 and L2 regularization were 0.1.

3. **FastText_CRF**: Similarly to WE_BERT_CRF, with the difference that *BERTweet* embeddings were replaced by *FastText* embeddings in the feature vector for each word. *FastText* vectors trained on similar diabetes-related tweets, which were well adapted to our use case.[9]

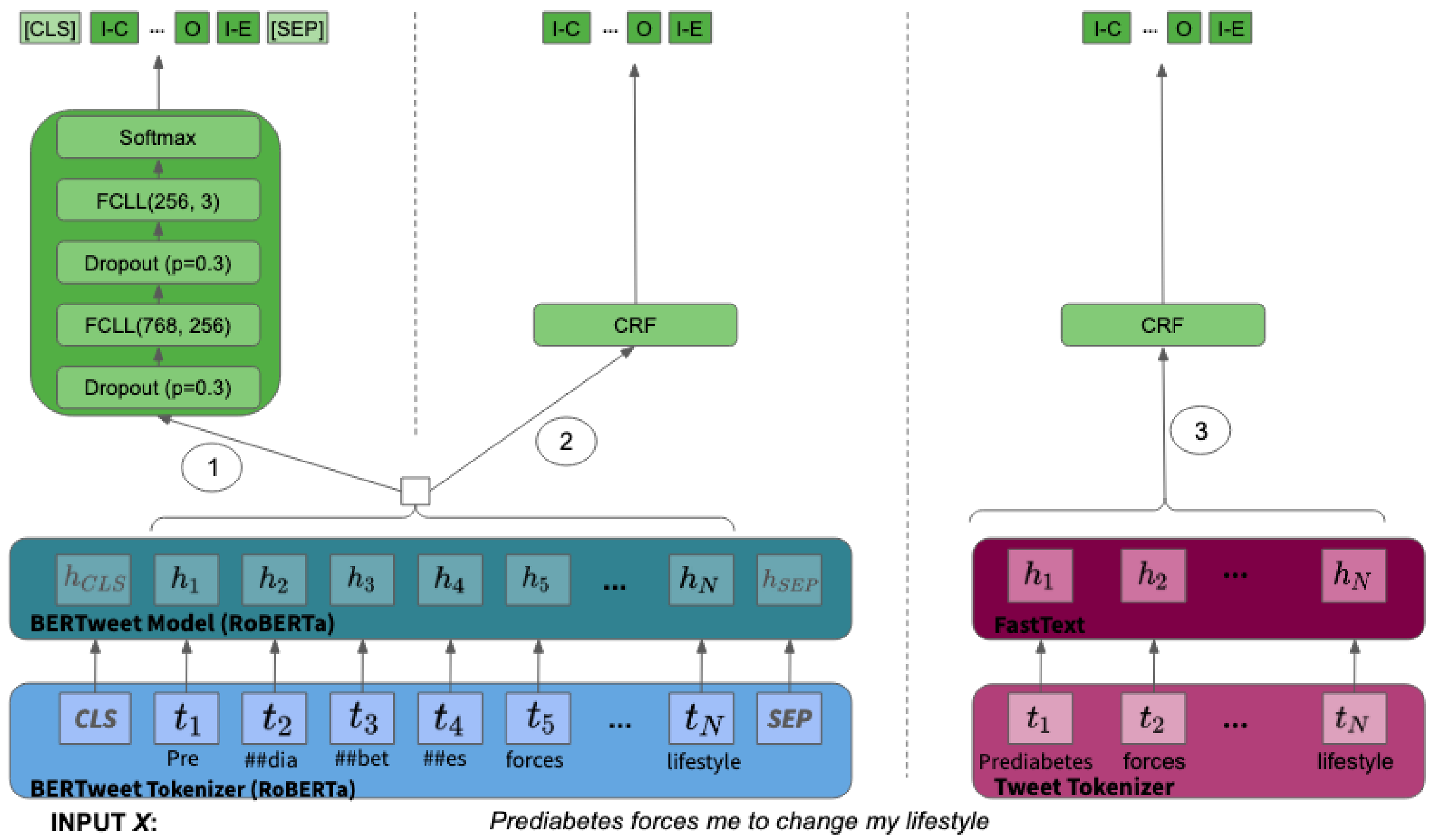

Figure 4: Model architectures - Cause-effect identification

## Clustering of causes and effects

A large part of *causes* and *effects* can be regrouped into similar concepts (clusters) to facilitate analyses and allow effective network analyses. We chose a semi-supervised, time-efficient approach in which 1,000 *causes* and 1,000 *effects* were randomly chosen and two researchers manually grouped these into clusters such as “diabetes”, “death”, “family”, “fear”, hereinafter referred to as “Parent clusters” to simplify understanding. The remaining *causes* and *effects* were then automatically compared to each element of all clusters, based on *FastText* vectors and cosine similarity, and associated to the cluster containing the most similar element. Experimentally a similarity threshold of 0.55 was determined; if a cause/effect had a similarity smaller than this threshold for all elements, a new cluster was created for this cause / effect.

These clusters were also visualised in an interactive cause-effect network, developed in D3, to enable further exploration of the cause-effect association about diabetes distress communication in social media

Python (version 3.8.8) and the deep learning framework pytorch (version 1.8.1) were used to implement the above-mentioned methods. The algorithms are open sourced under the following address: https://github.com/WDDS/Causal-associations-diabetes-twitter/

# Results

The following results were obtained from 482,583 sentences which were obtained from splitting the 562,013 personal, emotional, non-joke tweets into sentences; excluding questions; and including only sentences with more than 5 words.

## Model training and performance

### Causal sentences

The performances to detect causal sentences for the imbalanced dataset are illustrated in Table 2 for each round of the active learning loop, with each round having been trained on more data.

Highest accuracy was reached in round 4 with 71%. We applied the model of round 4 on all remaining tweets, as it was trained on the largest training data set, including difficult causal examples missed by earlier models, and is thus better at identifying complex causal sentences. The active learning strategy led us to increase the training data much quicker than without active learning and without loss in performance. This led to a clean database of 265,328 causal sentences with most noisy sentences removed.

| Round | N° sent. train | N° sent. test | Accuracy | Precision | Recall |
|---|---|---|---|---|---|
| 0 | 6,024 | 837 | 64.5 | 58.0 | 67.4 |
| 1 | 7,536 | 1,047 | 67.7 | 61.2 | 71.6 |
| 2 | 8,804 | 1,223 | 67.7 | 60.3 | 66.3 |
| 3 | 10,284 | 1,429 | 65.4 | 60.0 | 68.8 |
| 4 | 11,861 | 1,648 | **71.0** | 61.0 | 67.8 |

Table 2: Performance measures (macro) for each round of more training data

## Cause and Effect detection

After having identified the causal sentences, the cause-effect models were trained to extract the specific cause-effect pairs. The active learning strategy led to an extended dataset of 2,118 causal sentences, i.e. containing both cause and effect, of which 10% were used as a test set while the remaining 90% were further used to create a training and validation set with an 80:20% split. The performance of the different cause-effect models are listed in Table 3.

The best performing model was the CRF model with BERT embedding features (WE_BERT_CRF) with a precision, recall and F1 of 0.68. Surprisingly, it outperforms fine-tuning a BERT model, which is considered the gold standard of current NER tasks. A potential explanation for that is that BERT-based models make local decisions at every point of the sequence taking the neighboring words into account before its decision. In a situation like ours, with strong uncertainty on all elements, due to the complexity of the task, a single CRF layer model leveraging BERT features, making global decisions using the local context of each word, maximizes the probability of the whole sequence of decision better.
Moreover the CRF model with simpler FastText models achieved strong results as well with one reason being probably that the word embeddings were specifically trained on this diabetes corpus.

| **Models** | | **Prec** | **Rec** | **F1** |
|---|---|---|---|---|
| **BERT_FFL** | I-C | 0.48 | 0.46 | 0.47 |
| | I-E | 0.20 | 0.48 | 0.29 |
| | O | 0.91 | 0.77 | 0.83 |
| | macro | 0.53 | 0.57 | 0.53 |
| **WE_BERT_CRF** | I-C | 0.63 | 0.61 | 0.62 |
| | I-E | 0.49 | 0.49 | 0.49 |
| | O | 0.93 | 0.93 | 0.93 |
| | macro | 0.68 | 0.68 | **0.68** |
| **FastText_CRF** | I-C | 0.59 | 0.57 | 0.58 |
| | I-E | 0.45 | 0.38 | 0.41 |
| | O | 0.92 | 0.94 | 0.93 |

| | | | |
|---|---|---|---|
| macro | 0.65 | 0.63 | 0.64 |

Table 3: Performance measures for each of the four architectures

In consequence, the WE_BERT_CRF model was applied on all causal sentences leading to a dataset of 96,676 sentences with *cause* and associated *effect* predicted.

## Cause-effect description

The semi-supervised clustering led to 1,751 clusters. To remove noisy clusters through potential misclassifications, only clusters with a minimal number of 10 cause/effect occurrences were considered for the following analyses, resulting in 763 clusters. Note, the order of documents might affect the results, as different clusters might have been created. Please refer to Multimedia Appendix 4 for an overview over the 100 most largest clusters (automatically added clusters have “Other” as “Parent cluster”).

Table 4 provides an overview over the largest clusters, containing either cause or effect, on the left side and on the right side the most frequent cause-effect associations, excluding the largest cluster “Diabetes” as it will be studied separately. The cluster “Diabetes” is the largest one with 66,775 occurrences of “Diabetes” as either cause or effect (ex.: #diabetes, diabetes, diabetes mellitus) followed by “Death” with 16,989 (ex.: passed away, killed, died, suicide, etc.) and “Insulin” (ex.: insulin, insulin hormone, etc.) with 14,148. From the 30 largest clusters, 6 refer to nutrition, 4 to diabetes and 3 clusters to each of insulin, emotions and the healthcare system.

The most frequent cause-effect is “unable to afford insulin” which causes “death” expressed in 1,246 cases, followed by “insulin” causing “death” with 1,156 cases and “type 1 diabetes” causing “fear” with 1,054 cases.

| *Most frequent clusters* | | |
|---|---|---|
| **Parent cluster** | **cluster** | **N** |
| Diabetes | diabetes | 66,775 |
| Death | death | 16,989 |
| Insulin | insulin | 14,148 |

| *Most frequent cause-effect-relation (excluding cluster “diabetes”)* | | |
|---|---|---|
| **cause** | **effect** | **N** |
| unable to afford insulin | death | 1,246 |
| insulin | death | 1,156 |
| type 1 diabetes | fear | 1,054 |

| | | | | | | |
|---|---|---|---|---|---|---|
| Diabetes | type 1 diabetes | 11,693 | | type 1 diabetes | death | 999 |
| Emotions | fear | 10,160 | | rationing insulin | death | 805 |
| Glycemic variability | hypoglycemia | 9,547 | | type 1 diabetes | insulin | 751 |
| Symptoms | sick | 6,549 | | OGTT* | sick | 584 |
| Nutrition | overweight | 5,186 | | type 1 diabetes | hypoglycemia | 578 |
| Diabetes | type 2 diabetes | 4,909 | | insulin | hypo | 545 |
| Complications & comorbidities | neuropathy | 4,481 | | insulin | fear | 534 |
| Healthcare system | medication | 4,389 | | type 1 diabetes | insulin pump | 436 |
| Diabetes Technology | insulin pump | 4,307 | | finance | death | 423 |
| Nutrition | nutrition | 4,230 | | type 1 diabetes | sick | 400 |
| Emotions | anger | 4,149 | | insulin | sick | 385 |
| Health | OGTT* | 4,053 | | insulin | finance | 367 |
| Blood pressure | hypertension | 3,782 | | type 1 diabetes | anger | 356 |
| Healthcare system | finance | 3,767 | | insulin | medication | 305 |
| Nutrition | reduce weight | 3,589 | | insulin | anger | 296 |
| Insulin | unable to afford insulin | 3,381 | | OGTT* | fear | 293 |
| Nutrition | diet | 3,325 | | type 2 diabetes | death | 293 |
| Emotions | sadness | 3,153 | | type 2 diabetes | fear | 290 |
| Glycemic variability | hyperglycemia | 3,144 | | hypertension | death | 286 |
| Diabetes | suffer | 3,132 | | overweight | death | 280 |
| Diabetes Distress | depression | 2,810 | | type 1 diabetes | finance | 277 |
| Healthcare system | hospital | 2,721 | | hypoglycemia | insulin | 272 |
| Diabetes Distress | stress | 2,681 | | hypoglycemia | sick | 263 |
| Nutrition | sugar | 2,369 | | affordable insulin | death | 262 |
| Nutrition | fasting | 2,363 | | insulin | insulin pump | 255 |
| Insulin | rationing insulin | 2,244 | | complications | death | 248 |
| Health | gestational diabetes | 2,076 | | insulin | sadness | 240 |

Table 4: Left side column shows the most frequent clusters (causes and effects) with the number of occurrences. The last column shows the most frequent cause-effect relationships excluding the cluster “Diabetes”. *OGTT: Oral glucose tolerance test

The largest cluster “Diabetes” mainly occurs as a cause and its most frequent effects (“Death”, “fear”, “sick”) are visualised in Figure 5. From the 30 most numerous effects for “Diabetes”, 6 were related to “Nutrition” and 5 to “Complications & comorbidities” and 3 to each of “Diabetes distress”, “Emotions” and “Healthcare system”.

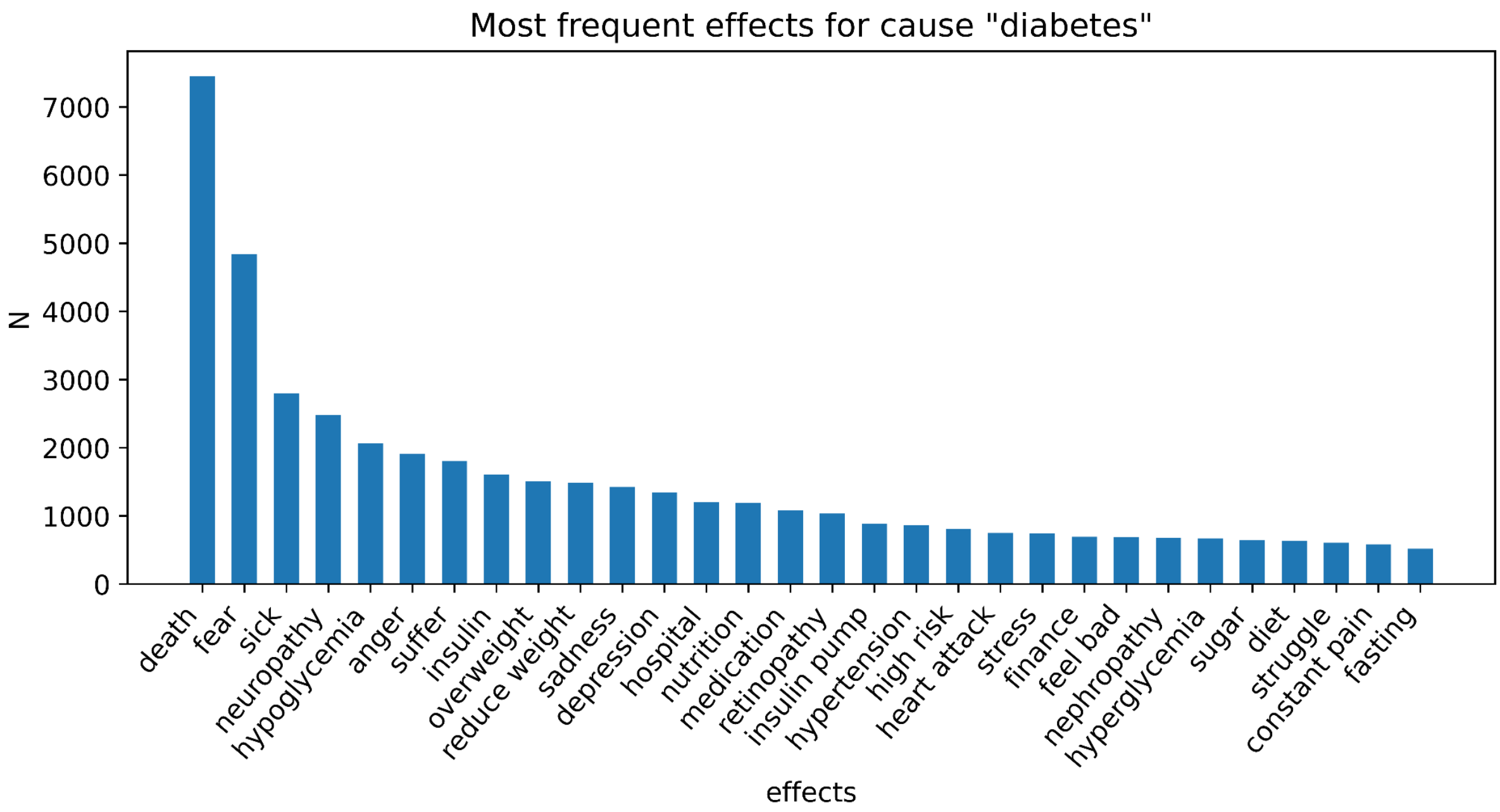

Figure 5: Most frequent effects for the largest cluster “Diabetes”

The interactive visualisation in D3 with filter options was published under https://observablehq.com/@adahne/cause-and-effect-associations-in-diabetes-related-tweets. We invite the interested reader to play with the graph to enhance understanding. Figure 6 provides an example graph of this visualisation showing only cause-effect relationships with at least 250 occurrences to ensure readability. It is striking that “death” seems to play such a central role as *effect* with various causes (“unable to afford insulin”, “rationing insulin”, “finance”, “insulin”, “Type 1 diabetes (T1D)”, “overweight”) pointing at it. Other central nodes are “Type 1 diabetes” acting as cause for “insulin pump”, “insulin”, “hypoglycemia (hypo)”, “sickness”, “finance” and emotions “anger” and “fear”, where latest has the strongest association; or the node “Insulin” mostly relating as cause to “sickness”, “medication”, “finance”, “death”, or “hypoglycemia” and “fear” and “anger”.

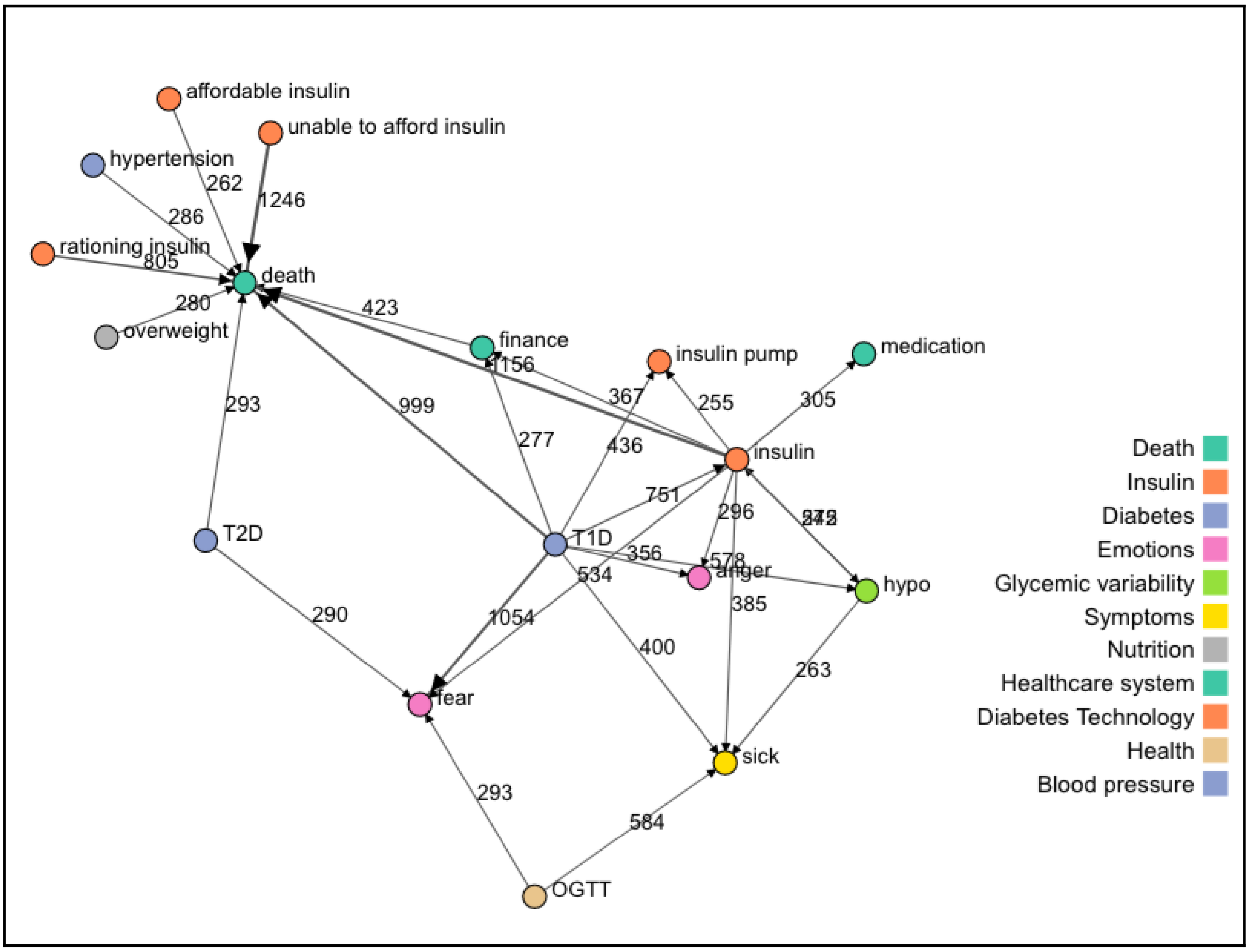

Figure 6: Cause-effect network with a minimum number of associations (edges) of 250. Accessible under: https://observablehq.com/@adahne/cause-and-effect-associations-in-diabetes-related-tweets

# Discussion

## Principal results

Our findings suggest that it is feasible to extract both explicit and implicit cause and associated effects from diabetes-related Twitter data. We demonstrated that by adopting the transfer learning paradigm and fine-tuning a pre-trained language model we were able to detect causal sentences. Moreover, we have shown that simply fine-tuning a BERT-based model does not always outperform more traditional methods such as relying on conditional random fields in the case of the cause-effect pair detection. The precision, recall and F1 numbers, given the challenging task and the imbalanced dataset, were satisfying. The semi-supervised clustering and interactive visualisation enabled us to identify “Diabetes” as the largest cluster acting mainly as the cause for "Death" and "fear". Besides, a central cluster was detected in "Death" acting as an effect for various causes related to insulin pricing, a link already detected in earlier works.[9] From a patients' perspective we were able to show that their main fear is insulin pricing expressed in the most frequent cause-effect relationship "unable to afford insulin" causing "death" or "rationing insulin" causing "death". As main diabetes distress related causes we identified fear from hypoglycemia, insulin, hypertension or the oral glucose tolerance test (OGTT).

## Comparison with the literature

Several former works have addressed causality on Twitter data. Doan et al. focused on three health-related concepts such as, "stress", "insomnia", "headache" as effects and identified causes using manually crafted patterns and rules.[14] However they only focused on explicit causality and excluded causes and effects encoded in hashtags and synonymous expressions.[14] On the contrary, we tackled both explicit and implicit causality, including causes and effects in hashtags, and exploiting synonymous expressions through the use of word embeddings. Kayesh et al. proposed an innovative approach, a novel technique based on neural networks which uses common sense background knowledge to enhance the feature set, but they focused on the simplified version of explicit causality in tweets.[18] Bollegala et al. developed a causality-sensitive approach for detecting adverse drug reactions from social media using lexical patterns and in consequence aiming at explicit causality.[39] Dasgupta et al. proposed one of the few deep learning approaches, due to the unavailability of appropriate training data, leveraging a recursive neural network architecture to detect cause-effect relations from text, but also only targeted explicit causality.[40] A Bert-based approach tackling both explicit and implicit causality is provided by Khetan et al. who used already existing labeled corpora not based on social media data.[23] Recently they further extended their work of explicit and implicit causality understanding in single and multiple sentence but in clinical notes.[41]

To the best of our knowledge, this is the first paper investigating both explicit and implicit cause-effect relationships on diabetes-related Twitter data.

## Strengths and Limitations

The present work demonstrates various strengths. First, by leveraging powerful language models we were able to identify a large number of tweets containing *cause-effect* relationships which enabled us to the detect cause-effect associations in 20% (96,676 / 482,583) of the sentences, contrary to other approaches which were able to identify causality in less than 2% of tweets.[14] Second, contrary to most previous work, we tackled both explicit and implicit *causal relationships*, an additional explanation for the higher number of *cause-effect* associations we obtained compared to other studies focusing only on explicit associations.[14] Third, relying fully on automatic machine learning algorithms avoided us from defining manually crafted patterns to detect causal associations. Fourth, operating on social media data that is expressed spontaneously and in real-time offers the opportunity to gain knowledge from an alternative data source and in particular from a patients' perspective, which might complement traditional epidemiological data sources.

A strong limitation is that *cause-effect* relations are expressed in tweets and this cannot be used for causal inference as the Twitter data source is uncertain and the information shared can be opinion or observation. Another shortcoming is that the performance of our algorithms to detect *cause-effect* pairs is not perfect. But the overall process and the vast amount of data minimizes this issue. The lack of recall is counterbalanced by the sheer amount of data and the lack of precision is counterbalanced by the clustering approach in which non-frequent causes or effects are discarded.[42] Labeling causes and effects in a dataset is a highly complicated task and we would like to emphasize that mislabelings in the dataset may occur. Enhancing data quality certainly is a strong point to address to further improve performance. The causal association structures learnt by the model from the training set, might not generalise completely when applied on the large amount of Twitter data. Besides, the active learning strategy certainly added noise to the model, as only positive samples were corrected, which could be improved in future investigations. Moreover, we would like to highlight that the diabetes related information shared on Twitter, may not be representative for all people with diabetes. For instance we observed a bigger cluster of causes/effect related to type 1 diabetes compared to type 2 diabetes, which is contrary to the real world.[43] A potential explanation for that is the age distribution of Twitter users.[44] But due to the large number of tweets analyzed, a significant variability in the tweets could be observed.

# Conclusion

In this work, we developed an innovative methodology to identify possible cause-effect relationships among diabetes-related tweets. This task was challenging due to addressing both explicit and implicit causality, multi-word entities, the fact that a word could be both cause or effect, the open domain of causes and effects, the biases occuring during labeling of causality, and the relatively small dataset for this complex task. We overcame these challenges by augmenting the small dataset via an active learning loop. The feasibility of our approach was demonstrated using modern *BERT*-based architectures in the preprocessing and causal sentence detection. A combination of *BERT* features and CRF layer were leveraged to extract causes and effects in diabetes-related tweets which were then aggregated to clusters in a semi-supervised approach. The visualisation of the cause-effect network based on Twitter data can deepen our understanding of diabetes, in a way of directly capturing patient-reported outcomes from a causal perspective. The fear of death due to the inability to afford insulin were main concerns expressed.

# Acknowledgements

None

# Competing Interests

None

# Funding

This work has been supported by the MSDAvenir Foundation, the French Speaking Diabetes Society and the Luxembourg Institute of Health. These study sponsors had no role in the design or the interpretation of the results of the present study. AA, FO and TC are supported by Epiconcept Company. Epiconcept was involved in the data collection and writing of the report. No study sponsor influenced the decision to submit the paper for publication.